\documentclass[conference]{IEEEtran}
\IEEEoverridecommandlockouts
\usepackage{cite}
\usepackage{amsmath,amssymb,amsfonts}
\usepackage{algorithmic}
\usepackage{graphicx}
\usepackage{textcomp}
\usepackage{xcolor}
\usepackage{float}
\usepackage[ruled,vlined]{algorithm2e}
\usepackage{booktabs}
\usepackage{multirow}
\usepackage{array}
\usepackage{tikz}
\usetikzlibrary{positioning, arrows}

\def\BibTeX{{\rm B\kern-.05em{\sc i\kern-.025em b}\kern-.08em
    T\kern-.1667em\lower.7ex\hbox{E}\kern-.125emX}}
\begin{document}


\title{Real-Time Loop Closure Detection in\\ Visual SLAM via NetVLAD and Faiss}
\author{\IEEEauthorblockN{Enguang Fan}
\IEEEauthorblockA{\textit{Department of Computer Science} \\
\textit{University of Illinois at Urbana-Champaign}\\
Urbana, IL, USA \\
enguang2@illinois.edu}
}




\maketitle

\begin{abstract}
Loop closure detection (LCD) is a core component of simultaneous localization and mapping (SLAM): it identifies revisited places and enables pose-graph constraints that correct accumulated drift. Classic bag-of-words approaches such as DBoW are efficient but often degrade under appearance change and perceptual aliasing. In parallel, deep learning-based visual place recognition (VPR) descriptors (e.g., NetVLAD and Transformer-based models) offer stronger robustness, but their computational cost is often viewed as a barrier to real-time SLAM.
In this paper, we empirically evaluate NetVLAD as an LCD module and compare it against DBoW on the KITTI dataset. We introduce a Fine-Grained Top-K precision--recall curve that better reflects LCD settings where a query may have zero or multiple valid matches. With Faiss-accelerated nearest-neighbor search, NetVLAD achieves real-time query speed while improving accuracy and robustness over DBoW, making it a practical drop-in alternative for LCD in SLAM.
\end{abstract}


\section{Introduction}
\begin{figure}[h]
    \centering
    \includegraphics[width=0.48\textwidth]{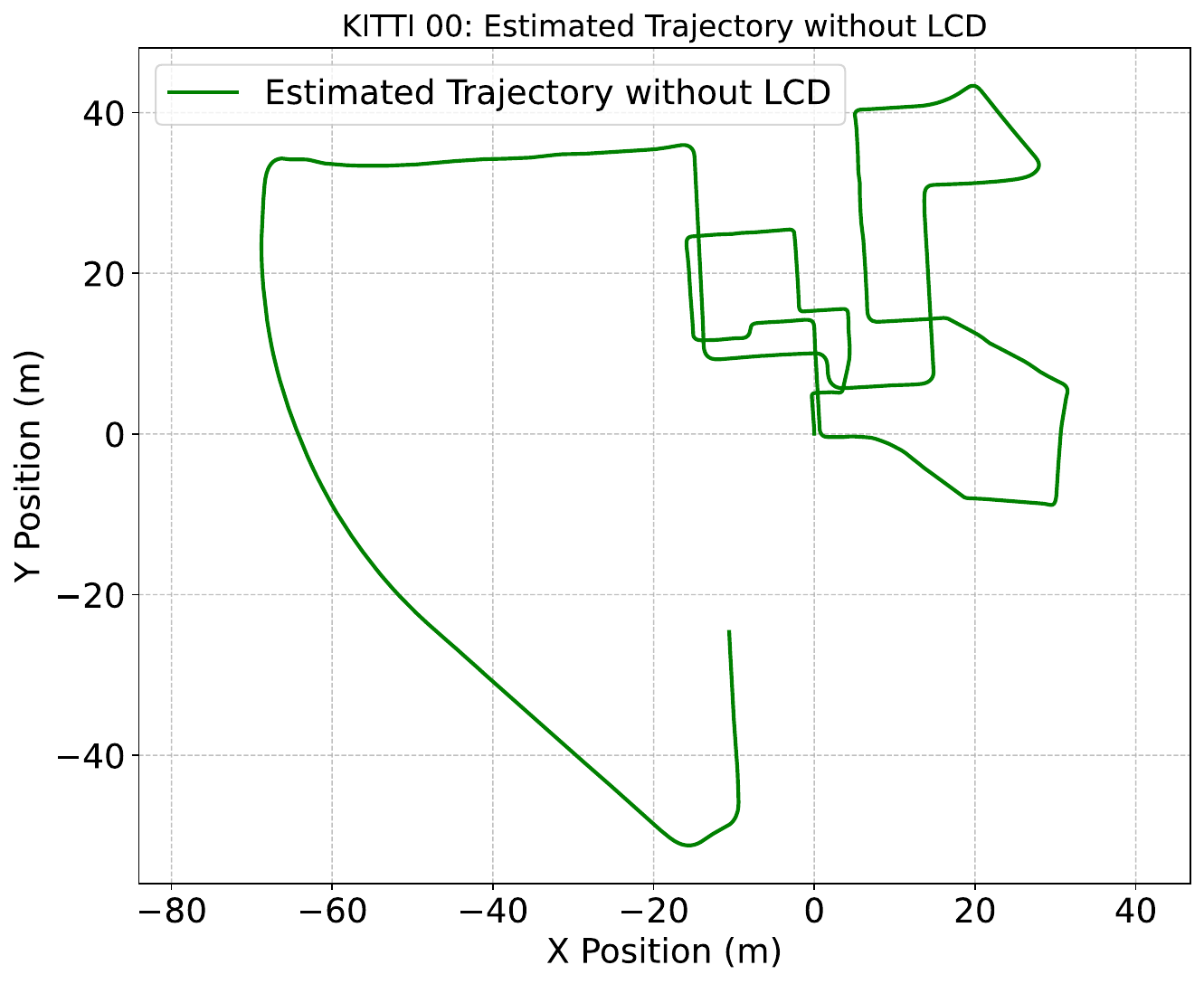}
    \caption{Monocular SLAM estimated trajectory for KITTI sequence 00 without Loop Closure Detection. Due to the lack of loop closure constraints, the estimated trajectory suffers from accumulated drift, leading to localization errors. Since this is a monocular SLAM system, the trajectory is only up to an unknown scale and the absolute meter values on the axes are not accurate.}
    \label{fig:no_lcd}  
\end{figure}

\begin{figure}[h]
    \centering
    \includegraphics[width=0.48\textwidth]{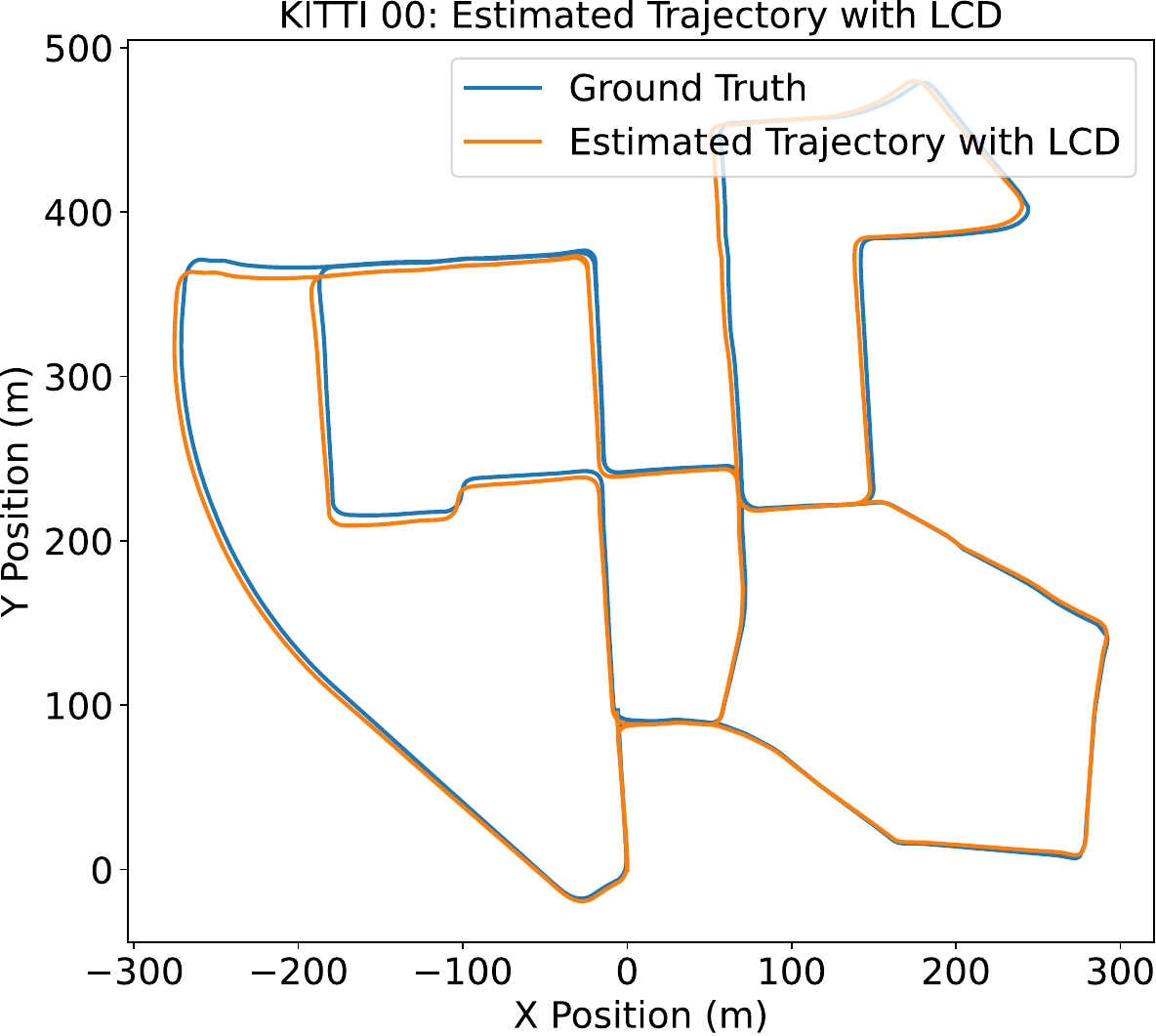}
    \caption{Monocular SLAM estimated trajectory and ground truth trajectory for KITTI sequence 00 with Loop Closure Detection. Since this is a monocular SLAM system, the estimated trajectory is only up to an unknown scale, and is scaled to best match the ground truth.}
    \label{fig:with_lcd}
\end{figure}

Loop Closure Detection (LCD) \cite{orb_slam, calc} is a fundamental component of Simultaneous Localization and Mapping (SLAM), enabling a robot to recognize previously visited locations and correct its estimated trajectory by eliminating accumulated drifts in its pose estimation. By identifying revisited places, LCD significantly enhances localization accuracy and ensures long-term SLAM consistency, especially in Monocular SLAM, where depth cannot be directly recovered from a single camera, leading to scale drift. Figure.~\ref{fig:no_lcd} illustrates the estimated trajectory of ORB-SLAM2 \cite{orb_slam2} for KITTI \cite{kitti} sequence 00 without a loop-closure detection module, where accumulated drift leads to a large deviation from the true path. In contrast, Figure.~\ref{fig:with_lcd} shows the corrected trajectory after applying loop closure detection, demonstrating how loop closures can constrain errors and maintain long-term localization accuracy. 

Traditional LCD methods, such as Dynamic Bag of Words (DBoW)~\cite{dbow}, rely on handcrafted local features (e.g., ORB~\cite{orb}, SIFT~\cite{sift}) and achieve high efficiency via fast feature extraction and an inverted-index search structure. A practical advantage is that SLAM pipelines can often reuse features already computed for visual odometry (frame-to-frame matching), so loop closure detection adds little extra feature-computation cost. The remaining overhead is the inverted-index lookup, which typically compares the query against only a subset of frames that share common visual words.
However, these methods suffer from limited robustness in dynamic or challenging environments and under extreme appearance changes, such as day/night transitions or seasonal variations. The reliance on local feature descriptors without spatial relationships also often leads to false loop closures due to perceptual aliasing, especially in repetitive, structured environments such as urban streets. A false positive example is shown in Figure.~\ref{fig:4537} and Figure.~\ref{fig:667}, which depict two frames retrieved by DBoW as loop closures despite belonging to different locations.

In contrast, deep learning-based Visual Place Recognition (VPR) methods, such as NetVLAD \cite{netvlad} and Transformer-based models (e.g., DINO v2 \cite{dinov2}, ViT-base \cite{vit}), have demonstrated higher accuracy by leveraging global feature embeddings learned from large-scale image datasets. These models extract more discriminative place descriptors, which implicitly capture geometric relationships, making them highly effective in challenging scenarios. However, the main drawback of deep learning-based approaches is their computational cost, as they require large neural networks and exhaustive nearest-neighbor searches for high-dimensional embedding vectors, making them impractical for real-time SLAM applications.

Moreover, while Visual Place Recognition (VPR) is closely related to Loop Closure Detection (LCD), they have fundamental differences in how matches are defined and handled. VPR typically assumes a one-to-one mapping between queries and database images~\cite{vpr-bench}, making high precision essential to avoid incorrect place recognition. In contrast, LCD operates within a SLAM framework, where:
\begin{itemize}
    \item A query frame \emph{may have no match} if it belongs to a trajectory that is not revisited.
    \item A query frame \emph{may have multiple valid matches} if it has multiple candidates that satisfy the relative pose difference constraint.
    \item False positives can be tolerated, as LCD systems can apply temporal and geometric consistency checks to reject incorrect matches.
\end{itemize}

Despite these differences, most deep-learning-based VPR methods have not been explicitly designed for LCD. Instead, they assume a static database and prioritize high-precision single-image retrieval. This paper provides a systematic empirical study of applying NetVLAD---a representative deep VPR descriptor---to LCD, assessing accuracy and query efficiency in a SLAM setting. In this paper, we empirically evaluate the effectiveness of NetVLAD as an alternative to DBoW for loop closure detection. Specifically, we aim to:

\begin{itemize}
    \item \textbf{Compare accuracy:} Compare NetVLAD and DBoW for loop closure detection, highlighting NetVLAD's robustness under appearance change and in dynamic scenes.
    \item \textbf{Assess efficiency:} Measure query efficiency with Faiss-accelerated nearest-neighbor search to evaluate real-time feasibility.
    \item \textbf{Evaluate with LCD-specific metrics:} Benchmark both methods using metrics tailored to LCD, including our Fine-Grained Top-K precision--recall curve.
\end{itemize}

\section{Related Work}
Loop Closure Detection (LCD) plays a crucial role in Simultaneous Localization and Mapping (SLAM) by enabling a robot to recognize previously visited locations and correct pose estimation drift. Traditional LCD approaches primarily rely on handcrafted visual feature descriptors, whereas recent advancements in deep learning-based Visual Place Recognition (VPR) methods have led to more robust, high-accuracy alternatives. However, the applicability of these deep learning models to real-time LCD has not been extensively studied. This section reviews both traditional LCD methods and modern deep-learning-based VPR techniques.
\subsection{Traditional Loop Closure Detection Methods}
Classic LCD methods primarily rely on handcrafted visual features, such as ORB \cite{orb}, SIFT  \cite{sift}, and SURF \cite{surf}, for detecting revisited locations. These features capture local keypoints and their descriptors, which are then used for image matching.

A widely adopted approach is Dynamic Bag of Words (DBoW) \cite{dbow}, which extends the traditional Bag of Words (BoW) paradigm by constructing a hierarchical vocabulary tree for efficient image retrieval. DBoW enables loop closure detection by:
Extracting local descriptors from images and associating them with pre-trained visual words.
Representing each image as a histogram of visual words, allowing for fast inverted index-based search.
Reusing feature descriptors from visual odometry, reducing computational overhead.
DBoW is highly efficient due to its compact representation and fast search structure. However, it suffers from several limitations:
Limited generalization in dynamic environments (e.g., day/night changes, seasonal variations).
Perceptual aliasing, where visually similar but distinct places produce false loop closures, particularly in structured environments like urban scenes.
Vocabulary dependency, requiring an offline-trained feature vocabulary that may not generalize to unseen environments.
Other traditional LCD methods, such as FAB-MAP \cite{fab_map}, introduce probabilistic models to improve place recognition reliability, but they still rely on handcrafted descriptors, making them vulnerable to appearance variations and scene dynamics.

\subsection{Deep Learning-Based Visual Place Recognition Methods}
Recent advancements in deep learning have led to the emergence of Visual Place Recognition (VPR) methods, which outperform traditional handcrafted approaches in challenging conditions. Unlike classic methods that rely on manually engineered features, deep learning-based VPR models learn discriminative feature embeddings from large-scale image datasets.

One of the most prominent deep-learning-based VPR methods is NetVLAD \cite{netvlad}, which enhances a CNN backbone with a VLAD-based \cite{vlad, all_about_vlad} pooling layer, generating global descriptors for place recognition. Unlike BoW-based methods, NetVLAD:
\begin{itemize}
    \item Learns feature representations directly from data, removing the need for predefined vocabularies.
    \item Captures global spatial relationships between features, reducing perceptual aliasing.
    \item Achieves higher accuracy than handcrafted descriptor-based methods.
\end{itemize}

Other notable deep-learning-based VPR approaches include:

\begin{itemize}
    \item Patch-NetVLAD \cite{patch_netvlad}, which incorporates localized descriptors to improve viewpoint robustness.
    \item Self-supervised learning methods like CALC  \cite{calc}, which trains an autoencoder to reconstruct HoG-based descriptors.
    \item Vision Transformers (ViTs) \cite{vit} and its variants, such as DINO \cite{dino} and DINOv2 \cite{dinov2}, which leverage self-attention mechanisms for feature learning and achieve state-of-the-art results.
\end{itemize}

Despite their success in VPR tasks, these deep learning methods have not been explicitly designed for LCD. Several challenges remain:

\begin{itemize}
    \item Computational cost: Methods like ViT and NetVLAD produce high-dimensional descriptors, requiring exhaustive nearest-neighbor searches, making real-time LCD infeasible without acceleration.
    \item Mismatch in retrieval criteria: While VPR typically prioritizes high precision (one correct match per query), LCD allows multiple correct matches based on trajectory constraints, requiring further post-processing steps.
\end{itemize}





\section{Methodology}
\label{method}

\begin{figure}[h]
    \centering
    \includegraphics[width=0.48\textwidth]{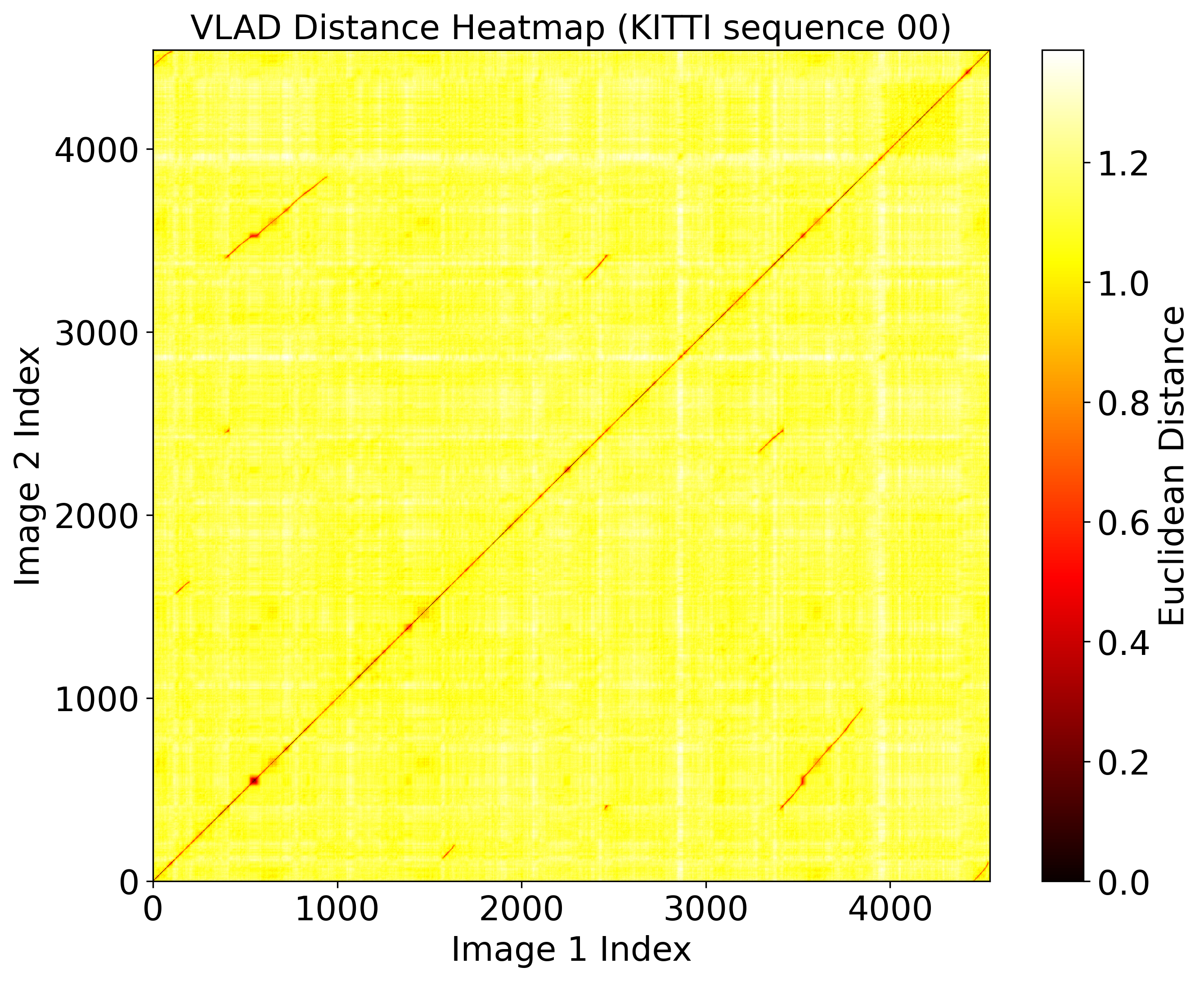}
    \caption{Similarity heatmap for KITTI sequence 00 based on NetVLAD Euclidean Distance. Darker regions indicate smaller distances (higher similarity) between frames. The diagonal represents self-to-self and adjacent frame matches, which are expected to have high similarity and should be ignored for loop closure detection.}
    \label{fig:heatmap_netvlad}
\end{figure}

\begin{figure}[h]
    \centering
    \includegraphics[width=0.48\textwidth]{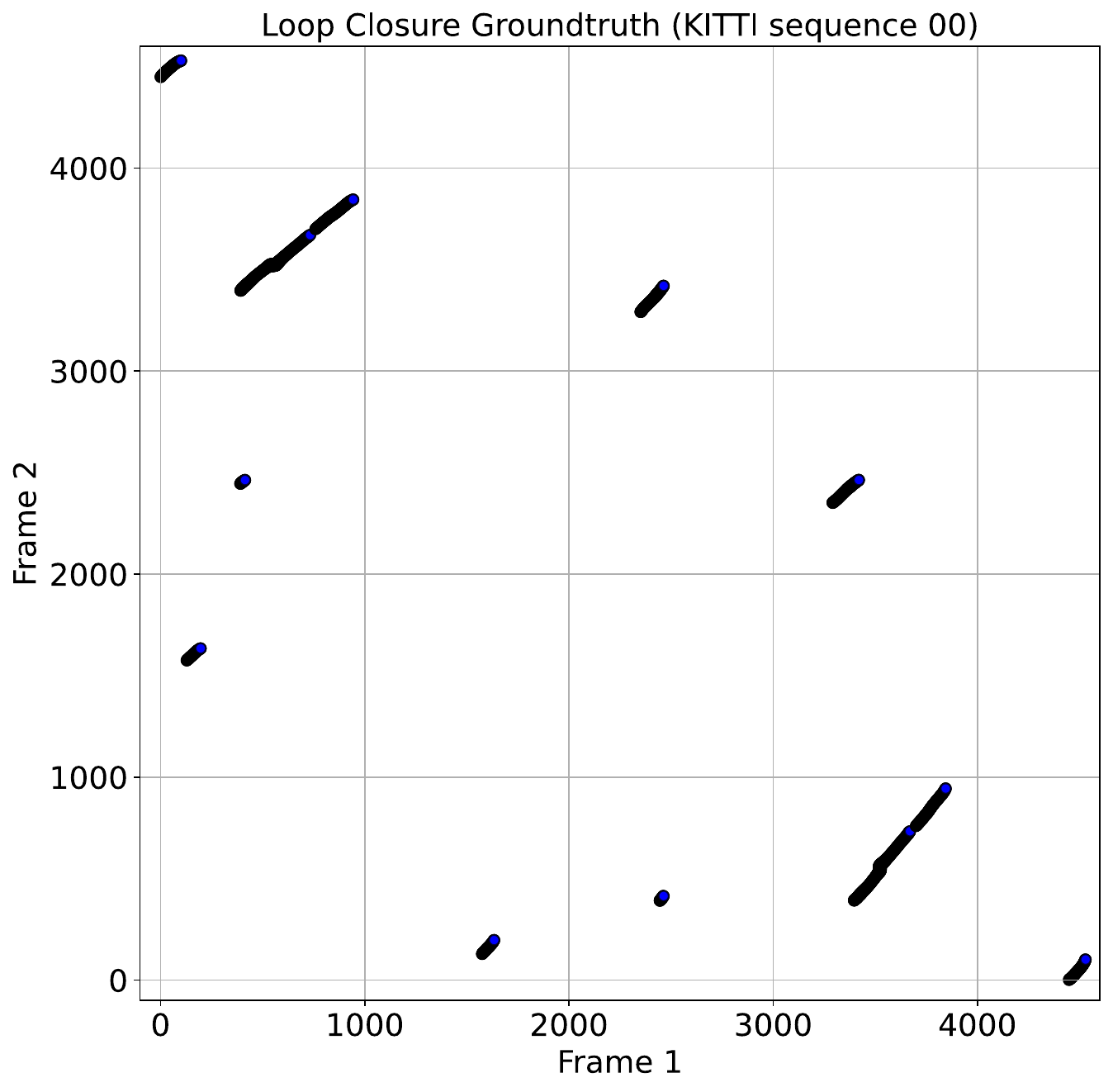}
    \caption{Pose ground truth clustering result for KITTI sequence 00. Frames with small relative pose differences form natural clusters.}
    \label{fig:groundtruth_clustering}
\end{figure}

In this section, we present our approach to applying Visual Place Recognition (VPR) techniques for the Loop Closure Detection (LCD) problem in SLAM. Our methodology replaces the traditional Dynamic Bag of Words (DBoW)-based LCD module with a deep-learning-based solution using NetVLAD. Unlike handcrafted feature-based methods, which struggle under environmental variations such as illumination changes and dynamic objects, our approach leverages learned global descriptors to improve robustness and accuracy while maintaining real-time feasibility. 
We visualize the frame-to-frame similarity scores of KITTI \cite{kitti} sequence 00 produced by NetVLAD in Figure.~\ref{fig:heatmap_netvlad}. This heatmap represents the Euclidean distance between descriptors of each frame, where darker regions indicate smaller distances (higher similarity). Additionally, Figure.~\ref{fig:groundtruth_clustering} shows the ground truth clustering results for loop closures, demonstrating the effectiveness of the defined pose constraints in identifying revisited locations.
We first introduce our SLAM pipeline, highlighting how NetVLAD is integrated as the loop-closure detection module. Next, we formally define the mathematical framework used to evaluate LCD performance. Finally, we propose a Fine-Grained Top-K precision--recall curve that reflects key differences between VPR and LCD evaluation. 
Our methodology aims to bridge the gap between accurate yet computationally expensive deep learning methods and efficient but less robust handcrafted approaches, making LCD both scalable and reliable for real-world SLAM applications.

\subsection{SLAM Pipeline with NetVLAD-based Loop Closure Detection}
\begin{figure*}[t]
    \centering
    \includegraphics[width=0.95\textwidth]{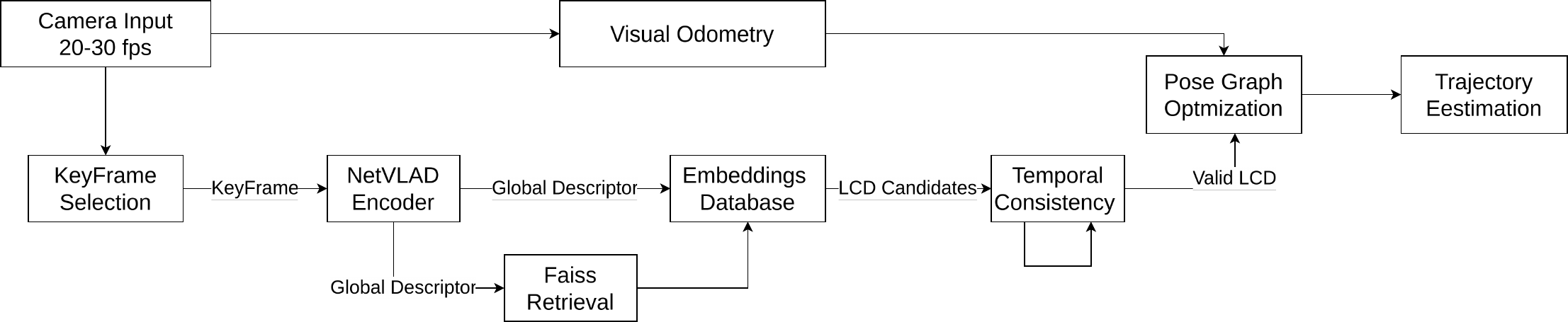}
    \caption{SLAM pipeline with NetVLAD-based Loop Closure Detection. The NetVLAD-based a loop-closure detection module replaces traditional DBoW-based methods, providing a more robust and accurate alternative while maintaining real-time feasibility through Faiss-based retrieval.}
    \label{fig:diagram}
\end{figure*}

By integrating NetVLAD as the LCD module, our SLAM pipeline achieves greater robustness to environmental variations while maintaining real-time feasibility through optimized retrieval using Faiss-based approximate nearest-neighbor search \cite{faiss}.

Figure~\ref{fig:diagram} illustrates our SLAM system with NetVLAD-based LCD. The pipeline consists of the following key modules:

\begin{itemize}
    \item \textbf{Visual Odometry (VO)}: Extracts local feature correspondences between consecutive frames to estimate incremental motion.
    \item \textbf{Keyframe Selection}: A subset of frames is dynamically selected to prevent redundant computations.
    \item \textbf{NetVLAD Encoder}: Each keyframe is processed by NetVLAD to generate a compact, high-dimensional embedding as its global descriptor.
    \item \textbf{Embeddings Database}: The extracted descriptors are stored in a database for efficient retrieval during loop closure detection.
    \item \textbf{Faiss-based Candidates Retrieval}: Matches a current keyframe against past keyframes using NetVLAD-derived global descriptors and retrieves top candidates using Faiss.
    \item \textbf{Temporal Consistency Check}: Instead of treating loop closure as a single-frame match, we enforce a sequence-to-sequence matching strategy. A valid loop closure should be supported by multiple consecutive keyframes forming a consistent group, rather than relying on isolated frame-to-frame matches. This helps reduce false positives caused by perceptual aliasing and ensures robustness against viewpoint changes.

    \item \textbf{Pose Graph Optimization}: Valid loop closures are used to correct accumulated drift in the estimated trajectory by applying pose graph optimization.
\end{itemize}

Compared to DBoW, our NetVLAD-based LCD provides the following advantages:

\begin{itemize}
    \item \textbf{Higher robustness}: NetVLAD learns feature embeddings that are invariant to viewpoint, illumination, and seasonal changes.
    \item \textbf{Improved accuracy}: Unlike DBoW, which relies on histogram-based visual word matching, NetVLAD retains spatial relationships in its learned representation.
    \item \textbf{Efficient retrieval}: Through Faiss-based nearest-neighbor search, we achieve query speeds comparable to DBoW while maintaining significantly higher accuracy.
\end{itemize}

By leveraging a deep-learning-based descriptor, our approach bridges the gap between traditional handcrafted methods and computationally expensive transformer-based models, making it a viable solution for LCD in SLAM. The proposed pipeline ensures a balance between speed and accuracy, making it suitable for real-world robotic applications.

\subsection{Mathematical Definitions}
We define the following terms to quantitatively analyze the LCD problem:

\begin{itemize}
    \item \textbf{Query Frame \(q\):} The current frame captured by the camera used to search for a matching previously visited frame. 

    \item \textbf{Query Retrieval Set \( R(q) \):} The returned query result is a set where each element is in the format \texttt{[frame ID, score]}:
    \begin{equation}
        R(q) = \{ (r_1, s_1), (r_2, s_2), \dots, (r_k, s_k) \}
    \end{equation}
    \begin{itemize}
        \item For DBoW, the score is based on a weighted visual word similarity measure.
        \item For deep-learning-based methods (e.g., NetVLAD, ViT), we define the matching score as a similarity value: cosine similarity, or the negative Euclidean distance between the global descriptors of the query and matched frames (so larger scores indicate closer matches).
        \item To ensure temporal consistency, we exclude temporally adjacent frames from being considered as valid loop closures, as they provide no additional information for correcting drift. This is also common practice in many SLAM systems \cite{orb_slam, orb_slam2, seqslam}.
    \end{itemize}

    \item \textbf{Ground Truth Matched Set \(G(q)\):} Defined as a set of frames that are valid loop closure matches for a given query. \begin{equation}
        G(q) = \{ g_1, g_2, \dots, g_m \}
    \end{equation} 
    where:
    \begin{itemize}
        \item \( G(q) = \emptyset \) if no loop closure exists for the query frame.
        \item \( |G(q)| = 1 \) if there is a unique revisited location.
        \item \( |G(q)| > 1 \) if multiple frames satisfy the relative pose difference threshold. 
    \end{itemize}
\end{itemize}

\subsection{True Positives, False Positives, False Negatives, and True Negatives}
Let \( \theta \) be the matching-score threshold. A retrieved frame \( r_i \) is accepted if \( s_i \geq \theta \), and rejected otherwise.
Using the above definitions, we classify retrieved results as follows:

\begin{itemize}
    \item \textbf{True Positives (TP)}: Retrieved frames that belong to the ground truth set and pass the threshold:
    \begin{equation}
        TP(q) = R(q) \cap G(q) \cap \{ r \mid s_r \geq \theta \}
    \end{equation}
    
    \item \textbf{False Positives (FP)}: Retrieved frames that are not in the ground truth but exceed the threshold:
    \begin{equation}
        FP(q) = R(q) \setminus G(q) \cap \{ r \mid s_r \geq \theta \}
    \end{equation}

    \item \textbf{False Negatives (FN)}: Ground truth frames that were not retrieved or had scores below the threshold:
    \begin{equation}
        FN(q) = G(q) \setminus R(q) \cup G(q) \cap \{ g \mid s_g < \theta \}
    \end{equation}

    \item \textbf{True Negatives (TN)}: Retrieved frames that are not in the ground truth set and were correctly rejected:
    \begin{equation}
        TN(q) = R(q) \setminus G(q) \cap \{ r \mid s_r < \theta \}
    \end{equation}
\end{itemize}

\subsection{Fine-Grained Top-K Precision-Recall Curve}
Existing VPR benchmarks use Precision-Recall (PR) curves based on \textbf{Top-1 retrieval}, assuming that there is always exactly one correct match per query~\cite{vpr-bench}. However, this assumption does not hold for LCD, where:

\begin{enumerate}
    \item A query may have multiple valid ground-truth matches, as any past frame within a certain relative pose threshold can be considered correct.
    \item A query may have no ground-truth match at all if it is not part of the revisited trajectory.
    \item False positives can be filtered out using temporal consistency and geometric consistency, unlike in VPR, where false positives directly impact accuracy.
\end{enumerate}

To better evaluate LCD performance, we propose the Fine-Grained Top-K precision--recall curve, which extends the standard PR curve by:
\begin{itemize}
    \item Considering each retrieved candidate individually, rather than treating the entire retrieval set as positive if it contains at least one correct match (as done in Recall@N).
    \item Evaluating the \textbf{Top-K retrieval} results, rather than just the Top-1 match, to fully utilize temporal and geometrical consistency filtering.
\end{itemize}

\noindent \textbf{Definition of Top-K Retrieval:}

Given a query frame \( q \), the retrieval set \( R(q) \) consists of the top \( K \) candidates ranked by their matching score:

\begin{equation}
    R_K(q) = \{ (r_1, s_1), (r_2, s_2), \dots, (r_K, s_K) \}
\end{equation}

We define precision and recall based on the Top-K retrieved results:

\begin{equation}
    \text{Precision@K} = \frac{|TP_K(Q)|}{|TP_K(Q)| + |FP_K(Q)|}
\end{equation}

\begin{equation}
    \text{Recall@K} = \frac{|TP_K(Q)|}{|TP_K(Q)| + |FN_K(Q)|}
\end{equation}

\( TP_K(Q) \), \( FP_K(Q) \) and \( FN_K(Q) \) represent the sets of true positives, false positives, and false negatives individually within the top \( K \) retrieved candidates, and \( Q \) is the set of all query frames. The Fine-Grained Top-K precision--recall curve is computed by varying threshold \( \theta \) across all possible values and plotting precision vs.\ recall, same as the standard PR curve.

\section{Evaluation}
\begin{figure}[h]
    \centering
    \includegraphics[width=0.48\textwidth]{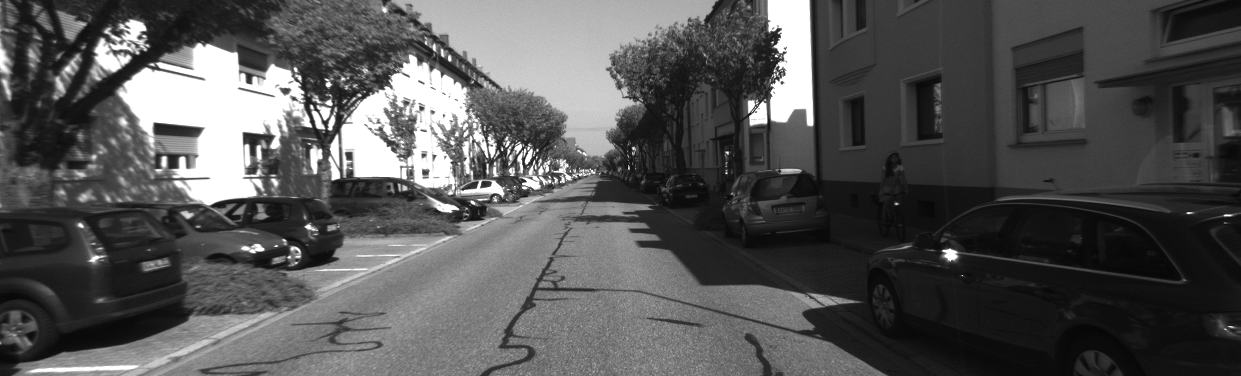}
    \caption{Frame 4537 from KITTI sequence 00}
    \label{fig:4537}  
\end{figure}

\begin{figure}[h]
    \centering
    \includegraphics[width=0.48\textwidth]{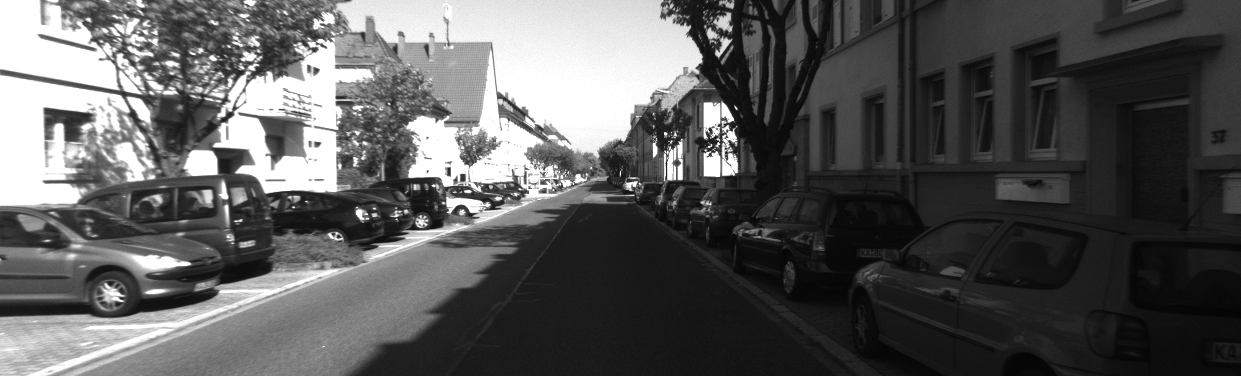}
    \caption{Frame 667 from KITTI sequence 00}
    \label{fig:667}  
\end{figure}

A major challenge in traditional loop closure detection methods like DBoW~\cite{dbow} is \emph{perceptual aliasing}, where visually similar but distinct places are incorrectly matched as loop closures. Figure~\ref{fig:4537} and Figure~\ref{fig:667} illustrate a false positive example where two frames are retrieved as a loop closure despite belonging to entirely different locations. This highlights a key limitation of manually crafted visual features, which struggle in structured environments such as urban streets, where repetitive features can lead to false positives. To systematically evaluate our method, we conduct experiments on a system equipped with an AMD Ryzen 7700X CPU and an Nvidia RTX 4070 Ti GPU.

\subsection{Experimental Setup and Metrics}
\subsubsection{Dataset: KITTI}

The KITTI~\cite{kitti} dataset is a widely used benchmark for evaluating visual odometry and SLAM systems. It provides camera streams and ground-truth poses for urban driving environments (via high-accuracy GPS/IMU). In our experiments, we use the left monocular grayscale camera stream at 20~Hz and precompute loop-closure ground truth. A loop closure between two frames is considered valid if their relative pose satisfies the following criteria:

\begin{itemize}
    \item The translational difference is below 1.5 meters.
    \item The rotational difference is below 0.3 radians.
\end{itemize}

To ensure meaningful loop closures, we exclude consecutive frames from being treated as valid matches, as they naturally appear similar but do not contribute additional constraints to pose graph optimization. To enforce this, we apply a naive temporal consistency check to exclude the most recent 100 frames.

\subsection{Accuracy Benchmarking Results}
\begin{figure}[h]
    \centering
    \includegraphics[width=0.48\textwidth]{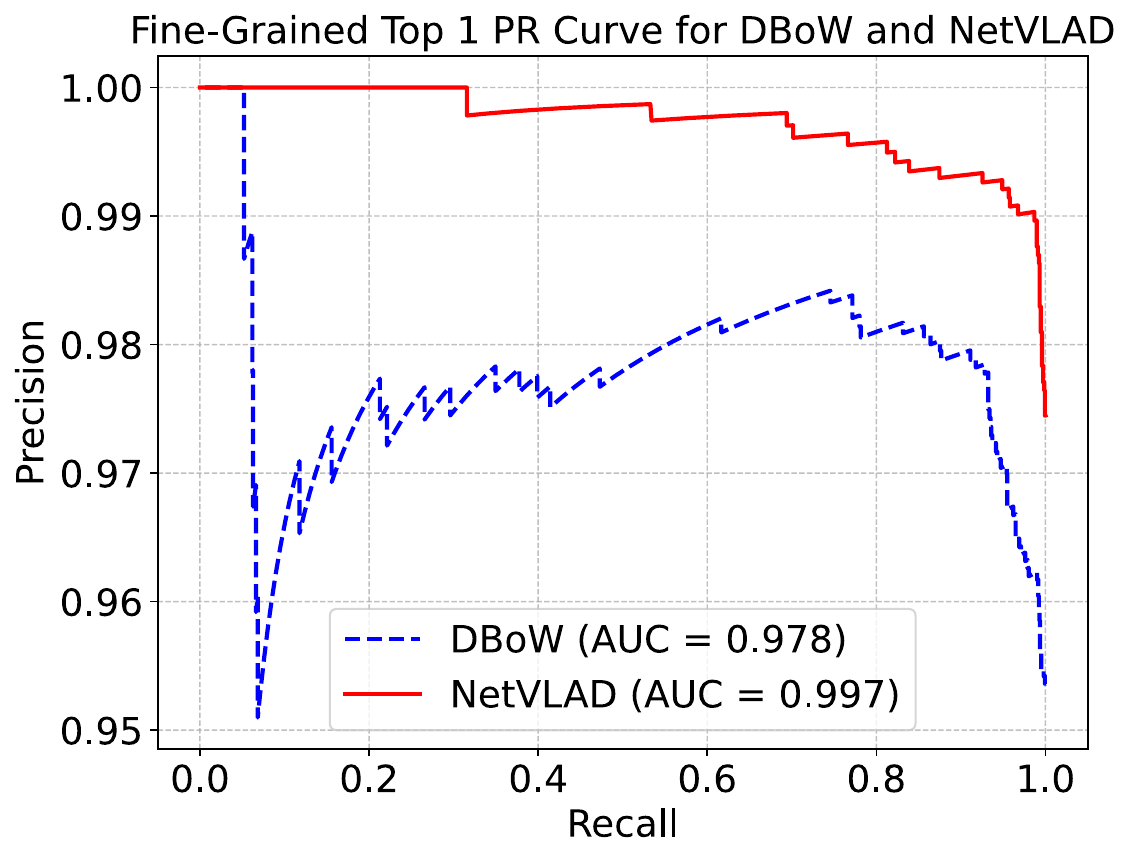}
    \caption{Fine-grained Top-1 PR curve for KITTI sequence 00}
    \label{fig:pr1}  
\end{figure}

\begin{figure}[h]
    \centering
    \includegraphics[width=0.48\textwidth]{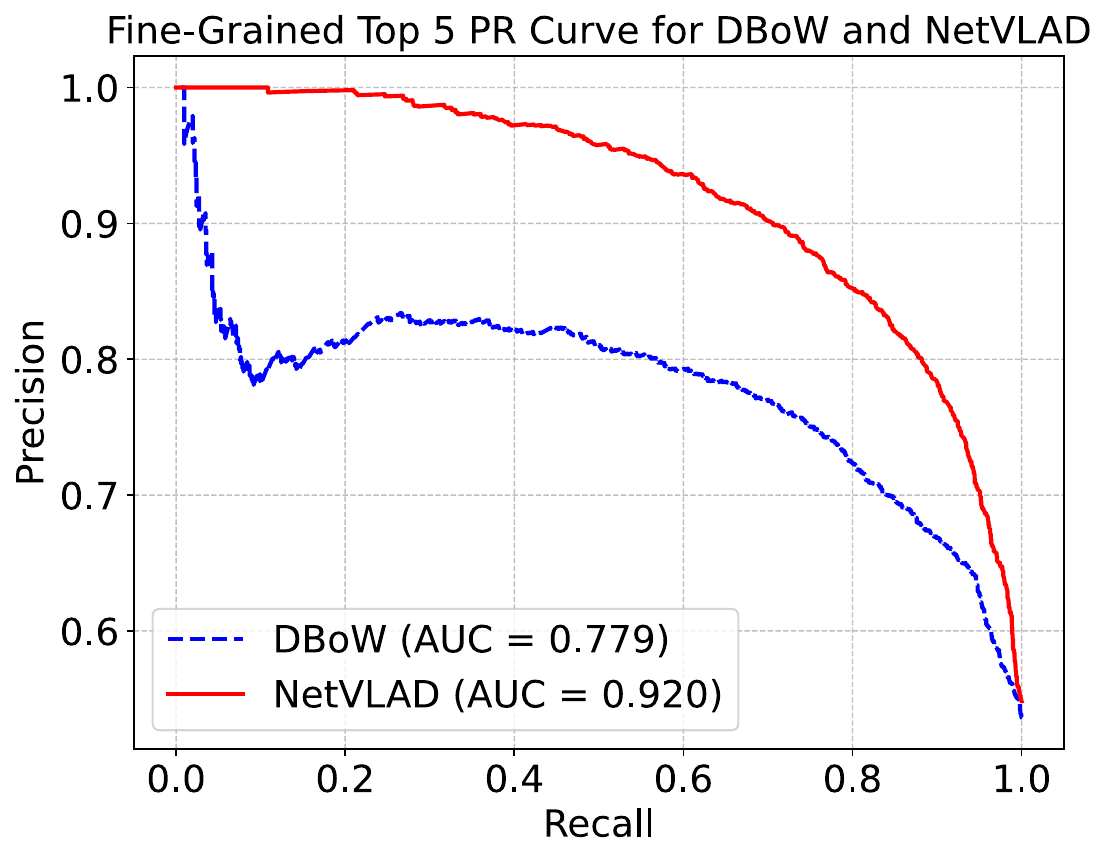}
    \caption{Fine-grained Top-5 PR curve for KITTI sequence 00}
    \label{fig:pr5}  
\end{figure}

\begin{figure}[h]
    \centering
    \includegraphics[width=0.48\textwidth]{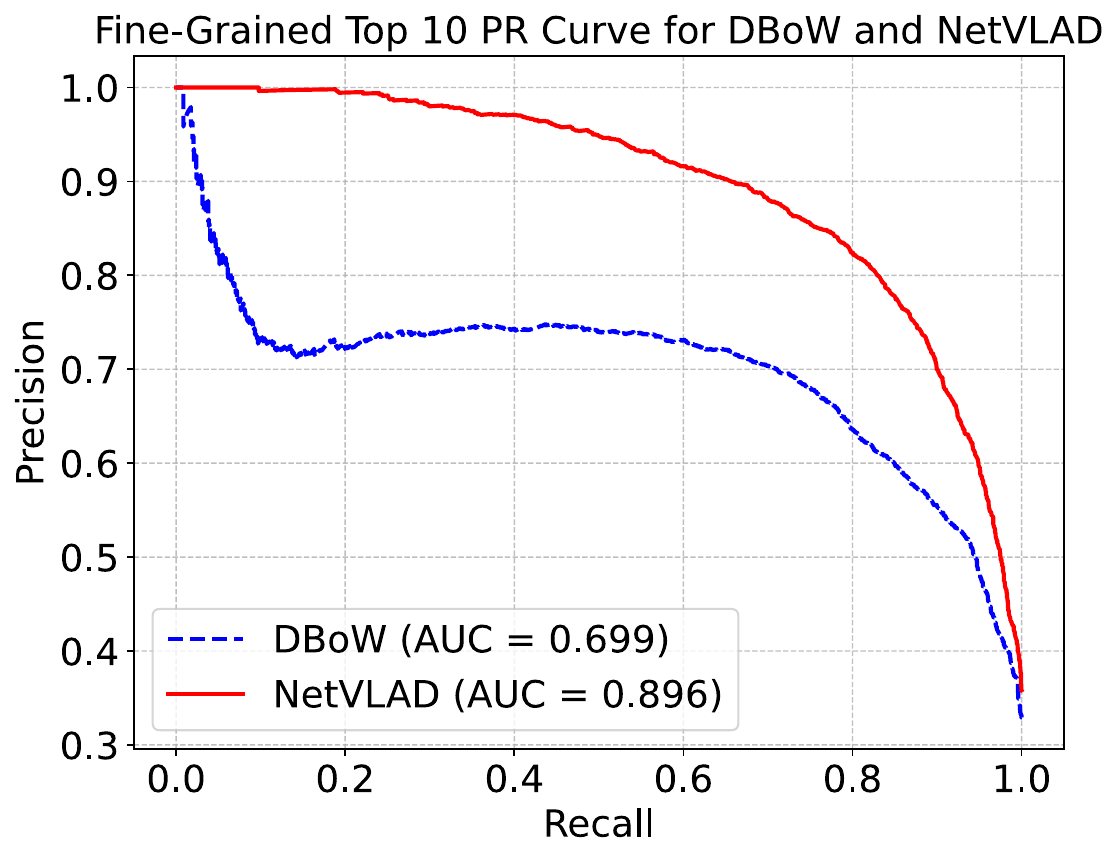}
    \caption{Fine-grained Top-10 PR curve for KITTI sequence 00}
    \label{fig:pr10}  
\end{figure}

\begin{figure}[h]
    \centering
    \includegraphics[width=0.48\textwidth]{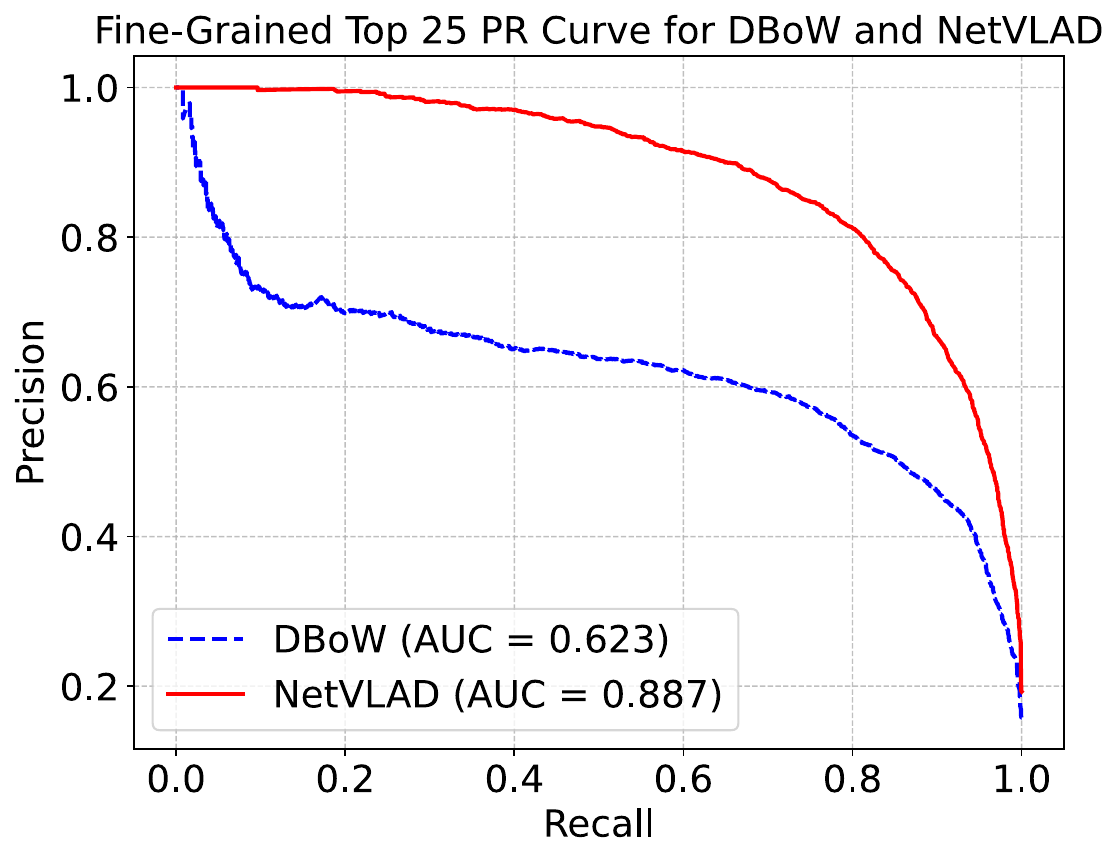}
    \caption{Fine-grained Top-25 PR curve for KITTI sequence 00}
    \label{fig:pr25}  
\end{figure}

We analyze the Fine-Grained Top-K Precision-Recall (PR) curves for both DBoW and NetVLAD under varying \textbf{Top-K} values \( \{1, 5, 10, 25\} \). Figure.~\ref{fig:pr1}--\ref{fig:pr25} illustrates how each method's precision and recall change as we adjust the number of top-ranked candidates for loop closure.

\begin{itemize}
    \item \textbf{Top-1:} 
    At this strict setting, both methods achieve relatively high precision but very low recall. Although the highest-scoring match is often correct, many genuine loop closures are overlooked. NetVLAD tends to maintain higher precision than DBoW due to more discriminative global descriptors.

    \item \textbf{Top-5 and Top-10:} 
    As \textbf{K} increases, recall improves because additional candidates are considered valid matches. However, precision drops slightly, since some incorrect matches inevitably make it into the top ranks. NetVLAD continues to outperform DBoW in terms of overall precision, reflecting its robustness in ranking true positives.

    \item \textbf{Top-25:} 
    At the highest \textbf{K}, the recall gains become more substantial. However, the drop in precision also becomes more pronounced, indicating a greater susceptibility to false positives. Even so, NetVLAD's precision remains higher across most recall levels, suggesting it is better at retrieving relevant loop closures from a larger candidate set.
\end{itemize}

Overall, these PR curves underscore the inherent trade-off between precision and recall: relaxing the criteria (larger \textbf{Top-K}) generally yields higher recall yet compromises precision. \textbf{NetVLAD} demonstrates a more favorable balance, consistently achieving higher precision while maintaining competitive recall. The optimal \textbf{Top-K} selection will depend on the specific requirements of the SLAM application, where some systems may prioritize precise loop closure detections and others may favor broader retrieval to maximize recall.

\subsection{Time Cost Analysis}

Query speed is another crucial factor in real-time SLAM applications. In a practical deployment scenario, a visual SLAM system operates on a robot with a camera frame rate of 20--30~FPS~\cite{kitti}. Since keyframes are typically selected every 3 to 4 frames, the LCD retrieval system must operate under a strict computational constraint: the total time for loop closure detection should be significantly lower than 100~ms per keyframe to maintain real-time performance, or equivalently, our system needs to process at least 10 keyframes per second.

The total query time can be further decomposed into two components: encoding time and retrieval time, expressed as:

\begin{equation}
T_{\text{query}} = T_{\text{encoding}} + T_{\text{retrieval}}
\end{equation}

where:
\begin{itemize}
    \item {Feature Encoding Time} (\(T_{\text{encoding}}\)) -- The time required to compute the image descriptor (e.g., NetVLAD descriptor extraction, BoW histogram generation).
    \item {Retrieval Time} (\(T_{\text{retrieval}}\)) -- The time required to search for matches in the reference database.
\end{itemize}

Let \( t_e \) represent the feature encoding time, and \( t_m \) represent the time required to match feature descriptors of two images as mentioned in \cite{vpr-bench}. The total query time can be expressed as \( t_R \), where:

\begin{equation}
t_R = t_e + O(N) \times t_m
\end{equation}

Here, \( O(N) \) represents the time complexity of the search algorithm for image matching, where \( N \) is the number of total images. 
\begin{itemize}
    \item For DBoW, query speed is significantly improved compared to naive linear search due to its inverted-index search, which typically compares a query against a subset of candidates that share common visual words. For evaluating all queries, the overall search complexity is:
    \begin{equation}
    O(M \cdot N)
    \end{equation}
    where \(M\) is the reduced candidate set per query, and \(M \ll N\). (Per query, the cost is \(O(M)\).)
    \item For NetVLAD, a naive linear scan across a database of size \(N\) has complexity:
\begin{equation}
O(N)
\end{equation}
per query (and \(O(N^2)\) if evaluated over all queries). This becomes costly at scale without acceleration. To address this, we integrate Faiss-based approximate nearest-neighbor search~\cite{faiss}, which uses low-level optimizations (e.g., SIMD) to greatly reduce retrieval time.
\end{itemize}

\subsection{Encoding and Retrieval Time Comparison}
We compare the feature encoding time and retrieval time between DBoW-based LCD and our NetVLAD-based LCD with Faiss on the KITTI \cite{kitti} sequence 00 dataset which contains 4541 images.

\subsubsection{Encoding and Retrieval Time Comparison}
Table~\ref{tab:encoding_retrieval_time} presents the total time required for encoding and retrieval for both methods, the time cost is for all 4541 images.

\begin{table}[h]
    \centering
    \caption{Comparison of Total Encoding and Retrieval Time}
    \label{tab:encoding_retrieval_time}
    \begin{tabular}{|c|c|c|}
        \hline
        \textbf{Method} & \textbf{Encoding Time (s)} & \textbf{Retrieval Time (s)} \\
        \hline
        ORB + DBoW & 19.028 & 9.389 \\
        NetVLAD + Faiss (Ours) & 42.600 & 2.870 \\
        \hline
    \end{tabular}
\end{table}

\subsubsection{Analysis of Results}  

\begin{itemize}
    \item \textbf{DBoW is faster} in computing image descriptors, requiring 19.028 seconds compared to 42.6 seconds for NetVLAD. This is expected, as DBoW relies on handcrafted feature extraction (ORB \cite{orb}, SIFT \cite{sift}), which is computationally lightweight and can be executed on the CPU with multithreading. NetVLAD, on the other hand, processes images through a deep CNN (VGG16 \cite{vgg16}) before applying VLAD \cite{vlad} pooling, making it more computationally expensive.

    \item \textbf{NetVLAD + Faiss is significantly faster} than DBoW for retrieval, requiring only 2.87 seconds compared to 9.389 seconds for DBoW. DBoW uses an inverted index search, which is efficient but still slower for large-scale datasets. NetVLAD, combined with Faiss, achieves retrieval speeds over 3$\times$ faster than DBoW, making it more suitable for real-time SLAM applications.
\end{itemize}
NetVLAD, despite having a higher embedding time than DBoW, achieves significantly faster retrieval speeds, making it fast enough for real-time SLAM. The average total processing time per image is 6.257 ms for DBoW and 10.013 ms for NetVLAD, both well below the 100 ms real-time constraint for loop closure detection. Given NetVLAD's superior accuracy and robustness to environmental variations, the trade-off in embedding time is justified, as it leads to more reliable loop closure detection and improved SLAM performance, while still remaining viable for real-time applications.

\section{Conclusion}
In this paper, we investigated the applicability of deep-learning-based Visual Place Recognition (VPR) methods---specifically NetVLAD---to Loop Closure Detection (LCD) in SLAM. Our comparative experiments with Dynamic Bag of Words (DBoW) on the KITTI dataset demonstrated three key insights:

\begin{itemize}
    \item \textbf{Accuracy and Robustness:} NetVLAD consistently outperformed DBoW across multiple Top-K settings, achieving higher precision and recall in dynamic and visually repetitive urban scenes. Its learned global descriptors mitigated perceptual aliasing issues that often plague handcrafted feature-based methods.

    \item \textbf{Fine-Grained Evaluation:} By introducing a Fine-Grained Top-K Precision-Recall Curve tailored for LCD, we highlighted the fundamental differences between VPR (single-match assumption) and LCD (multiple valid matches). Our results showed that NetVLAD maintained superior performance at both low and high Top-K values.

    \item \textbf{Real-Time Feasibility:} Despite the higher-dimensional descriptors produced by deep learning, our integration of Faiss-based nearest-neighbor search enabled total query speeds competitive with DBoW. This indicates that deep-learning-based approaches can be optimized to operate under the strict time constraints required by real-time SLAM.

\end{itemize}

Overall, our findings confirm that NetVLAD is a viable and effective alternative to DBoW for LCD in SLAM, offering robustness against changing environmental conditions while still meeting practical runtime requirements. Future work will explore further optimization of the deep descriptors (e.g., model compression, GPU acceleration) and extend our approach to other datasets and sensor modalities. We also plan to investigate more advanced temporal and geometric consistency checks to further reduce false positives. 

\bibliographystyle{IEEEtran}
\bibliography{thesisrefs}
\end{document}